# Large Language Models Meet Graph Neural Networks for Text-Numeric Graph Reasoning

Haoran Song[1]*, Jiarui Feng[2]*, Guangfu Li[3], Michael Province[4,5], Philip Payne[1], Yixin Chen[2], Fuhai Li[1,6#]

[1]Institute for Informatics, Data Science and Biostatistics (I2DB), [2]Department of Computer Science and Engineering, [4]Division of Statistical Genomics, [5]Department of Genetics, [6]Department of Pediatrics, Washington University School of Medicine, Washington University in St. Louis, St. Louis, MO, USA.
[3]Department of Surgery, School of Medicine, University of Connecticut, CT, 06032, USA.
*Co-first authors; [#]Correspondence: Fuhai.Li@wustl.edu

**Abstract**

In real-world scientific discovery, human beings always make use of the accumulated prior knowledge with imagination pick select one or a few most promising hypotheses from large and noisy data analysis results. In this study, we introduce a new type of graph structure, the text-numeric graph (TNG), which is defined as graph entities and associations have both text-attributed information and numeric information. The TNG is an ideal data structure model for novel scientific discovery via graph reasoning because it integrates human-understandable textual annotations or prior knowledge, with numeric values that represent the observed or activation levels of graph entities or associations in different samples. Together both the textual information and numeric values determine the importance of graph entities and associations in graph reasoning for novel scientific knowledge discovery. We further propose integrating large language models (LLMs) and graph neural networks (GNNs) to analyze the TNGs for graph understanding and reasoning. To demonstrate the utility, we generated the text-omic(numeric) signaling graphs (TOSG), as one type of TNGs, in which all graphs have the same entities, associations and annotations, but have sample-specific entity numeric (omic) values using single cell RNAseq (scRNAseq) datasets of different diseases. We proposed joint LLM-GNN models for key entity mining and signaling pathway mining on the TOSGs. The evaluation results showed the LLM-GNN and TNGs models significantly improve classification accuracy and network inference. In conclusion, the TNGs and joint LLM-GNN models are important approaches for scientific discovery.

## Introduction

**The problem: the need of text-numeric graph (TNG) for real-world scientific discovery.** In the real world, many complex systems can be represented using interconnected graph-structured data, such as knowledge graphs, cell signaling graphs, and chemical structure graphs. Graph AI, based on graph neural networks (GNN) models, is one effective approach for graph understanding and reasoning. A major limitation of the current existing graph-based reasoning is that they are developed for either text-attributed[1–4] or numeric-attributed[5–7] graphs. It is not consistent with and not efficient for the real-world scientific knowledge discoveries. In real-world scientific discovery, human beings always make use of the accumulated prior knowledge with imagination pick select one or a few most promising hypotheses from large and noisy data analysis results. Therefore, the prior knowledge and numeric data should be integrated during scientific discovery and knowledge reasoning. Herein, we introduce a new type of graph structure, the text-numeric graph (TNG), which is defined as graph entities and associations have both text-attributed information and numeric information. The TNG is an ideal data structure model for novel scientific discovery via graph reasoning because it integrates human-understandable textual annotations or prior knowledge, with numeric values that represent the observed or activation levels of graph entities or associations in different samples. Together both the textual information and numeric values determine the importance of graph entities and associations in graph reasoning for novel scientific knowledge discovery. We further propose integrating large language models (LLMs) and graph neural networks (GNNs) to analyze the TNGs for graph understanding and reasoning.

**Application scenarios of LLM-GNN for TNG:** To demonstrate the utility, we generated the text-omic(numeric) signaling graphs (TOSG), as one type of TNGs, in which all graphs have the same entities, associations and annotations, but have sample-specific entity numeric (omic) values using single cell RNAseq (scRNAseq) datasets of different diseases. We proposed joint LLM-GNN models for key entity mining and signaling pathway mining on the TOSGs. The evaluation results showed the LLM-GNN and TNGs models significantly improve classification accuracy and network inference. There are rich prior knowledge of life science entities, like proteins, drugs, diseases and phenotypes[1]. Also the textual prior knowledge are available from LLM genrations, like using ChatGPT 40-mini: "APOE4 (Apolipoprotein E4) is an allele (genetic variant) of the APOE gene, which encodes a protein involved in lipid (fat) metabolism. APOE is essential for the transport of cholesterol and other lipids in the bloodstream, and it plays a key role in the central nervous system (CNS), particularly in the brain. …", "APOE2 (Apolipoprotein E2) is one of the three major isoforms of the APOE gene, with APOE3 and APOE4 being the other two. Unlike APOE4, which is associated with an increased risk of Alzheimer's disease and other health issues, APOE2 has a somewhat protective role in terms of brain health and other aspects of lipid metabolism….", and **"**mTOR (mechanistic Target of Rapamycin) is a crucial protein kinase that regulates various cellular processes involved in growth, metabolism, and homeostasis. It serves as a central regulator of cell growth, proliferation, protein synthesis, autophagy (the process by which cells degrade and recycle their components), and metabolism, responding to a variety of internal and external signals such as nutrients, energy status, growth factors, and stress. …", and 'KRAS (Kirsten Rat Sarcoma viral oncogene homolog) is one of the most commonly mutated genes in pancreatic ductal adenocarcinoma **(**PDAC**)**, which is the most common and aggressive form of pancreatic cancer**.** KRAS mutations are present in approximately 90% of PDAC cases, making it a central driver of the disease. The mutation of KRAS in PDAC is considered one of the earliest and most critical events in its development and progression."

On the other hand, single-cell RNA sequencing (scRNA-seq) is a powerful genomic tool used to analyze the transcriptomes of individual cells, offering high-resolution insights into gene expression at the single-cell level[8–11]. In recent years, scRNA-seq has gained significant popularity due to its ability to profile gene expression patterns and characterize cellular heterogeneity with unprecedented precision. This technique enables the investigation of diverse biological processes and molecular mechanisms by identifying differentially expressed genes across distinct cell populations and subpopulations. By revealing genes that are upregulated or downregulated under specific conditions, scRNA-seq provides a deeper understanding of cellular responses. Moreover, scRNA-seq holds great promise in uncovering intricate communication networks within cellular microenvironments, facilitating the study of intercellular interactions and niche-specific signaling pathways.

A set of computational approaches have been introduced to extract active signaling pathways and intracellular communication networks using scRNA-seq data. These methods typically fall into two categories. The first involves applying statistical frameworks, such as correlation, regression, or Bayesian models, to interpret cellular interaction[12]. For example, CellPhoneDB[13] is designed to map ligand-receptor interactions between different cell types, while CCCExplorer[14] expands on this by uncovering not only ligand-receptor pairs but also downstream signaling pathways through the analysis of differentially expressed genes. NicheNet[15] takes this a step further by integrating various interaction databases and employing a predictive regression model to estimate the potential impact of ligands on downstream target genes. CytoTalk[16], on the other hand, uses the Steiner tree algorithm to infer novel signal transduction networks based on gene co-expression data. The second type of method applies advanced deep learning methods [17,18,19,20,21,22,23,24,18,25]. to automatically extract useful information based on model prediction. One representative method is PathFinder[26]. In PathFinder, they first extract extensive biological meaningful paths from the gene-gene interaction database. Next, they design a novel graph transformer model to predict the condition of the individual cell based on scRNA-seq data and extracted paths. Finally, the model will output important paths based on the attention value of each path to form the intra-cell communication networks.

The unprecedented performance of Large Language Models (LLMs) inspired many researchers to leverage them for solving problems in the biomedical field. scInterpreter[27] leverages LLMs to directly interpret scRNA-seq data for cell type annotation. They first extract descriptions for the top 2048 genes and use a text embedding model to project it to vector space. Next, the projected embedding along with the prompt with be input into LLMs for answering questions. Some other methods including scGPT[28] and scBERT[29] utilize the architecture and training schema of LLMs to train the model directly on large-scale single-cell datasets for cell-type annotation, and multi-batch integration. We refer readers to a recent survey[30] for further discussion. However, there is no existing method that can leverage both the transcriptomic information from scRNA-seq data and textual information from existing literature yet.

Herein, we introduce the new type of graph, i.e., text-numeric graph (TNG), and LLM-GNN for TNG graph reasoning. We applied the joint LLM-GNN models for analyzing the text-omic(numeric) signaling graphs (TOSG), one type of TNG, to mine key signaling targets and signaling pathways using single cell RNAseq (scRNAseq) datasets. The evaluation results showed the LLM-GNN models significantly improve classification accuracy and network inference. The details of methods and results are in the following sections.

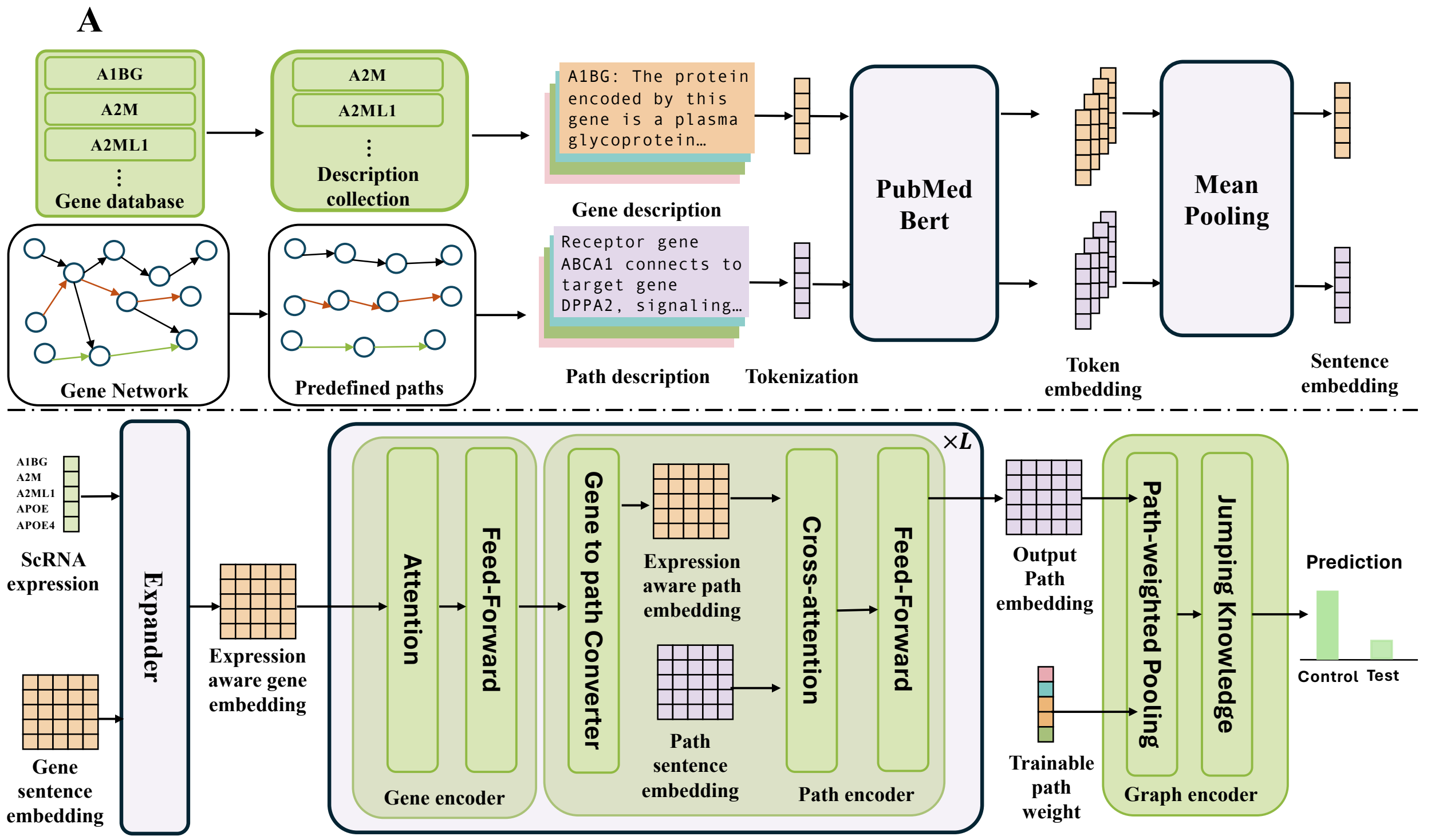

**Figure 1. (A)** Model architecture overview of LLM-GNN on Text-omic(numeric) graphs.

## Methodology

### Notations

A gene graph is denoted as $G = (V, E)$, where $V$ is the set of gene nodes with $|V| = n$, $E$ is the set of edges and $E \subseteq V \times V$. The node embedding set is denoted by $X = [x_1, x_2, \dots, x_n]^T \in R^{n \times d}$, where $x_u \in R^d$ is the embedding vector of the node $u$. The graph structure is defined by an adjacency matrix $A \in [0,1]^{n \times n}$, where $A_{uv} = 1$ indicate there is an edge from the node $u$ to node $v$ and $A_{uv} = 0$ otherwise. Further, a set of paths sampled from a graph is denoted as $P = \{p_1, p_2, \dots, p_p\}$, where $p_m$ is the $m$-th path, which is a list to store the nodes of the path in order. Paths can have different lengths, and we denote the length of path $m$ be $l_m$.

### The proposed LLM-GNN on TNG models.

The overview of the proposed model is shown in **Figure 1**. The model consists of three parts: gene encoder, path encoder, and graph encoder. For the gene and path encoder, we add sentence embedding generated from LLMs to obtain additional knowledge about genes and paths from the training corpus of the LLMs. The sentence embedding will be combined with gene expression for the prediction of the cell condition and output important paths. In the following, we discuss each module respectively.

### Sentence embedding from LLM

To acquire knowledge about genes and paths from LLM, we utilize LLM to generate sentence embedding for genes and paths. Specifically, for each gene/protein, the textual prior knowledge is available from the four datasets NCBI Gene, GeneCards, UniProtKB/Swiss-Prot, and Tocris. In addition, we screened these genes through the network to generate gene pairs with biological connections, and converted them into text, such as "In this pathway: receptor gene ABCA1 connect to target gene DPPA2." We sent the converted text to PubMedBert 18 to generate Path-Embedding. The description contains the basic information about genes like their functionality and bioprocess. Next, the description will pass to a well-trained LLM to get the final embedding for each token in the sentence. To obtain the sentence embedding, we use the mean pooling to average the embedding of all tokens in the sentence. For each path, we use the natural language to describe the path. Here is an example: "In this pathway: receptor gene ABCA1 connects to target gene DPPA2, signaling gene DPPA2 connects to signaling gene FXR2." Similarly, we use the procedure to convert the description to the sentence embedding. We denote the text description of gene $v$ as $t_v$ and sentence embedding as $se_v \in R^{d_{llm}}$, where the $d_{llm}$ is the hidden size of the LLM. Similarly, we denote the text description of path $m$ as $t_m$ and sentence embedding as $se_m \in R^{d_{llm}}$. The equation to generate sentence embedding is shown as the follow:

$$se_v = mean\big(LLM(t_v)\big),$$

$$se_m = mean\big(LLM(t_m)\big),$$

Where $LLM$ denote the LLM and $mean$ denote the mean pooling. For LLM, we choose the PubMedBert, which is pre-trained on large-scale PubMed papers. It has superior knowledge in the biomedical domain. In this way, we can make the most of the information on LLM to help the next module understand the relationship between genes and their connections.

**Gene encoder**

This model is a graph-based neural network for processing gene expression data. It combines gene expression data, sentence embedding from LLM, and node connectivity and spatial location information to produce a gene feature map understood by our model. The architecture of the gene encoder is a transformer-based model, which stacks $L$ transformer layer. Additionally, we add inductive bias of gene network structure, we add centrality encoding, spatial encoding, and edge encoding. The input to the gene encoder contains two parts: scRNA expression and the sentence embedding of genes. Let $ge_v \in R^1$ be the gene expression of gene $v$. First, the gene expression is integrated with the sentence embedding of gene $v$ through the Expander module. In Expander, the gene expression is first expanded by Multi-Layer Perceptrons (MLPs) and then concatenated with the sentence embedding. Finally, another MLP is used to reduce the hidden size of the concatenated embedding to the model's hidden size. The equation is shown as the follows:

$$x_v = MLP\left(Concat\big(se_v, MLP(ge_v)\big)\right),$$

Where $Concat$ is the concatenation function to combine multiple vector into one single large vector, $x_v \in R^{h_{emb}}$ and $d$ is the hidden size of the model. Next, the processed embedding of all genes is stacked and input to the transformer layer in the gene encoder.

Let $H^l \in R^{n \times h_{emb}}$ denotes the stacked embedding of all genes at gene encoder layer $l$, and $H_v^l \in R^{h_{emb}}$ be the embedding of gene $v$ and we have $x_v = H_v^0$. Each gene encoder layer contains a multi-head self-attention module and a point-wise feed-forward network (FFN) with residual connection applied between each part. The computation of multi-head self-attention is :

$$Q^{l,i} = H^{l-1}W_Q^{l,i}, \qquad K^l = H^{l-1}W_K^{l,i}, \qquad V^l = H^{l-1}W_V^{l,i},$$

$$head_i = Attention\big(Q^{l,i}, K^{l,i}, V^{l,i}\big) = SoftMax\left(\frac{Q^{l,i}{K^{l,i}}^T}{\sqrt{d_k}}\right)V^{l,i},$$

$$O^l = Concat(head_1, \dots, head_h)W_O^l,$$

where $W_Q^{l,i}$, $W_K^{l,i}$, $W_V^{l,i} \in R^{h_{emb} \times d_k}$, and $W_O^l \in R^{hd_k \times h_{emb}}$ are all trainable weight matrix, $h$ is the number of heads, $O^l \in R^{n \times h_{emb}}$ is the output from the multi-head self-attention in layer $l$. However, the vanilla transformer cannot be used directly on the graph structure data as it lacks a critical part for encoding the topological information into the model. To deal with this issue, we leverage and modify the method of Graphormer. Specifically, we introduced centrality encoding, spatial encoding, and edge encoding. The centrality encoding is used to embed the graph centrality information into the model. Given the input data $X$, the computation of centrality encoding is:

$$H^0 = X + Z^-\{deg^-(G)\} + Z^+\{deg^+(G)\},$$

where the $Z^-, Z^+$ are all trainable embedding vectors and $deg^-(G), deg^+(G): G \rightarrow R^n$ are the function to compute the in-degree and out-degree of each node in the graph $G$. The spatial and edge encoding is used to encode the graph structure into the model. With the spatial and edge encoding, the self-attention is revised as:

$$head_i = SoftMax\left(\frac{Q^{l,i}K^{l,i^T}}{\sqrt{d_k}} + b_i\{\phi(G)\} + c_i\right)V^{l,i},$$

where $b^i$ is trainable embedding vectors to encode the spatial information at head $i$ and $\phi(G): G \rightarrow R^{n\times n}$ is the function to compute the spatial information. In the proposed model, as each gene is a node in graph and each gene has its unique identity, we directly use it as the spatial encoding function. Specifically, for each gene, we add a unique node ID, and each ID corresponds to a trainable ID embedding. We denote the ID embedding of all genes as $I \in R^{n\times h_{emb}}$. Then, the spatial function $\phi(G) = II^T$. $c^i \in R^{n\times n}$ is the edge embedding and $c^i_{uv} = x_{uv}$, where $x_{uv}$ is the edge feature of the edge between node $u$ and node $v$. For node pair without edge, we add a special edge type to it.

Next, the output $O^l$ will then be fed into a point-wise feed-forward network. The computation of the point-wise feed-forward network is:

$$FFN(x) = ReLu(xW^l_1 + b^l_1)W^l_2 + b^l_2,$$

where $W^l_1 \in R^{h_{emb}\times h_{emb}}$, $W^l_2 \in R^{h_{emb}\times h_{emb}}$, $b^l_1 \in R^{h_{emb}}$, and $b^l_2 \in R^{h_{emb}}$ are all trainable weight matrix and bias. The output is fed into the next layer of the gene encoder for further processing.

**Path encoder**

This module further converts gene embeddings into path embeddings by connecting each layer of the gene encoder to the path encoder layer, intending to learn path-level information. In short, it is like encoding the interactions of genes and their roles in specific biological pathways, thereby helping the model understand the specific functions and action pathways of these genes in organisms. Each layer of gene encoder also connect to a path encoder layer. The output gene embedding from the layer are further convert to path embedding and to learn path-level information. Given the output embedding in the graph and the pre-defined path list of the graph.

The details of the pre-defined path list are illustrated below. Suppose there are $p$ unique paths in the path list $P$, where the length of the $m$-th path is $l_m$ and the total number of nodes in the path list is $k$ (count repeated nodes in different paths). Denote the node embedding output from the layer $l$ as $H^l$, we first learn a path-specific embedding through:

$$U_i^l = scatter(H^l)_i W_u^l + b_u^l,$$

where $W_u^l \in R^{h_{emb} \times u}$ and $b_u^l \in R^u$ are all trainable weight matrix, $scatter$ is a function to reorder and scatter the node in the graph into the order of the pre-defined path list. For example, suppose there are 5 embedding genes output from the node encoder. That is $H^l \in R^{5 \times h_{emb}}$. We label each gene from 1 to 5. Suppose there are two paths. The first path is 1->3->4. The second path is 2->3->4->5. Then, the $scatter(H^l)$ will output a new matrix with the size of 7 and each row represents a gene in a path. For instance, the 3rd row is $H_4^l$ since it is the 3rd gene in the first path. $U^l \in R^{k \times u}$ is the learned path-specific embedding. For convenience, we denote $U_{m,i}^l$ as the embedding of $i$-th node in the $m$-th path. Then, path positional encoding and path edge encoding are introduced to encode additional information for all paths. Let $\bar{U}^l$ be the result embedding after the special encodings. We have:

$$\bar{U}_{m,i}^l = U_{m,i}^l + p_i^l + e_{i,i+1}^l,$$

Where $p_i^l$ is the learnable positional encoding vector and its value only depends on the position $i$, $e_{i,i+1}^l$ is the learnable edge encoding to encode the edge type between $i$-th node and $i+1$-th node. Next, the score of each node within the path is computed by:

$$S_{m,i}^l = tanh(\bar{U}_{m,i}^l W_{s1}^l + b_{s1}^l) W_{s2}^l + b_{s2}^l$$

$$\bar{S}^l = ScatterSoftMax(S^l),$$

where the $W_{s1}^l \in R^{u \times r}$, $b_{s1}^l \in R^r$, $W_{s2}^l \in R^{r \times r}$, and $b_{s2}^l \in R^r$ are all trainable parameters. $ScatterSoftmax$ is the softmax function working within each path. The $S^l \in R^{k \times r}$ is the final $r$ set important score for each node in each path. We let the $r \times u = h_{emb}$ for simplicity. After we obtain the $S^l$, the expression aware path embedding is computed by:

$$P^l = Flatten\left(ScatterSum\left(S^l * \bar{U}^l\right)\right)$$

The $*$ is the point-wise product working on each set of important scores. That is, for each set of important scores, we do a point-wise product of that set of scores and $U^l$, which results in total $r$ sets. The $ScatterSum$ function is the summation on each path. The $Flatten$ is the function to flatten the embedding of all sets.

Finally, the sentence embedding of the path is used and integrated with the expression aware path embedding the cross-attention module. Specifically, the computation in the cross-attention module is similar to the self-attention. However, the key and value tensor will be replaced with path sentence embedding and the query tensor is the expression aware path embedding.

**Graph encoder**

Finally, the graph encoder is used to make a prediction and also extract important paths. The graph encoder consists of two parts, the first part is a trainable path weight and the sigmoid function to assign each path with different scores. The second part is the jumping knowledge network to combine the graph embedding in each layer and compute the final embedding.

In the model, the graph embedding is learned by integrating all the path embedding from each layer, which requires an important score for each path. Normally, the score is computed based on one sample. However, such a score is not robust and may vary a lot even given a minor variation of the path embedding[31,32]. To avoid the issue and learn a robust important score across the whole dataset, the trainable path score $M \in R^p$ is introduced. $M$ is identical to all samples and layers and learned through backpropagation. The path important score is computed by:

$$I = Sigmoid(M),$$

where $I \in R^p$ is the important score for each path. Next, the graph embedding of layer $l$ is computed by:

$$g^l = IP^l,$$

where $g^{l}$ is the graph embedding of layer $l$. The final step of the graph encoder is to integrate the graph embedding of each layer and learn a final embedding. Here we utilize the idea of JumpingKnowledge network[33] and compute the final graph embedding by:

$$G = MaxPooling\left(Concat(g^1, g^2, \dots, g^L)\right),$$

where $MaxPooling$ is the max pooling function and $G \in R^{h_{emb}}$ is the final graph embedding learned by the PathFinder. Finally, the graph embedding is used to classify the cell sample into the corresponding condition (control/test). The prediction is a typical binary prediction computed by:

$$p = SoftMax(GW_p),$$

Where $W_p \in R^{h_{emb} \times 2}$ is the trainable projection matrix and $p$ is the predicted distribution.

**Cell importance**

We defined Cell Importance as a comprehensive indicator of cell frequency changes and gene expression differences between disease and non-disease states:

$$\text{Cell Importance} = \alpha \times \left(\left|\mathrm{f}_{\text{diseased}} - \mathrm{f}_{\text{healthy}}\right|\right) + \sum_{j=1}^{m} \left|e_{\text{diseased}}^{(j)} - e_{\text{healthy}}^{(j)}\right|$$

Specifically, Cell Importance is calculated using the following formula: where α=0.5 is used to balance the contribution of frequency changes and gene expression differences, $\left|f_{\text{diseased}} - f_{\text{healthy}}\right|$ represents the sum of the differences in cell numbers between disease and non-disease states, reflecting changes in cell frequency; $\sum_{j=1}^{m} \left|e_{\text{diseased}}^{(j)} - e_{\text{healthy}}^{(j)}\right|$ represents the difference in gene expression, with j denoting individual genes within a single cell, capturing gene expression changes between disease and non-disease states based on scRNA-seq data.

**Trajectory analysis**

We propose a cell trajectory analysis pipeline based on the minimum spanning tree (MST) and pruning method. First, we use the model to generate embeddings for cell populations, combining scRNA-seq data and text embedding data of genes and pathways generated by LLM to obtain embedding vectors for each cell population. Then, we calculate the cosine distance between cell populations, defined as:

$$d_{ij} = 1 - \frac{\mathbf{v_i} \cdot \mathbf{v_j}}{|\mathbf{v_i}||\mathbf{v_j}|}$$

In this context, $v_i$ and $v_j$ represent the embedding vectors of cell populations i and j, respectively, while $|v_i|$ is the norm of the vector, and $d_{ij}$ denotes the cosine distance between cell populations i and j. Next, we construct a Minimum Spanning Tree (MST) among these cell populations to find the shortest loop-free path connecting all cell populations. The objective is to find the Minimum Spanning Tree $T \subseteq G$, whose edge set $E_T$ satisfies:

$$\sum_{(i,j)\in E_T} d_{ij} = \min\left(\sum_{(i,j)\in E'} d_{ij}\right)$$

Among them, $E_T$ is the edge set $T$, and E′ is the edge set of a possible spanning tree. Finally, we perform the pruning step, where we filter out paths that do not meet the time flow requirements based on the time annotations. Specifically, we retain only the paths that flow from a previous time point to the target cell, while filtering out paths with incorrect time flow direction. Given a target cell $\mathrm{C_t}$, and $\mathrm{C_j}$ represents an intermediate cell along the path to the target cell, we only keep the paths that satisfy the time flow condition, expressed as:

$$E_{\text{pruned}} = \left\{ (i,j) \in E_T \mid \text{Time}(C_i) < \text{Time}(C_j) \text{ for } C_j = C_t \right\}$$

Through this process, we ultimately obtained a cell trajectory network that conforms to time flow, represented as:

$$T_{\text{final}} = (V, E_{\text{pruned}})$$

, where $T_{\text{final}}$ denotes the final cell trajectory network that adheres to the time flow constraint. It is represented by the graph $(V, E_{\text{pruned}})$ where $V$ the set of nodes (cells) $E_{\text{pruned}}$ is the set of edges that only include paths satisfying the time flow condition.

## Datasets

### scRNA-seq data of human cirrhosis disease

The cirrhosis single-cell RNA sequencing dataset is collected from the GEO database, accession number GSE13610316, and will be used to generate pathways and compare with the benchmark for validation. It includes non-parenchymal cells collected from 5 healthy people and 5 patients with cirrhosis. After processing, the single-cell data had a total of 59,854 non-parenchymal single cells. The data were aligned to the GRCh38 and mm10 (Ensembl84) reference genomes as required and processed using the 10X Genomics Cell Ranger v.2.1.0 single-cell software suite to estimate unique molecular identifiers (UMIs). Genes expressed in less than 3 cells, cells expressing less than 300 genes, or cells with mitochondrial gene content exceeding 30% of the total UMI were excluded from the analysis. We isolated three major study cells, including 6197 Endothelial cells, 9173 Macrophages, and 20950 T cells.

### scRNA-seq data of Alzheimer's disease cohort on mice

Alzheimer's disease scRNA-seq data with the ID number GSE164507 were collected from the Gene Expression Omnibus (GEO) database 17. The raw data were processed using the Seurat R package and followed the previous research methods 17. Specifically, we selected cell samples under two different conditions, called TAFE4_tam and TAFE4_oil. TAFE4_tam refers to mice with APOE4 gene knocked out from microglia, while TAFE4_oil refers to mice with APOE4 gene present. Specifically, we collected excitatory neurons (Ex), microglia (Mic), and astrocytes (Ast) from the TAFE4 group from the dataset, with sample numbers of 13,604, 3,874, and 734, respectively.

**scRNA-seq data from human of PDAC**

The scRNA-seq dataset of PDAC tumors included samples collected from 31 patients who underwent standard treatments. The experiments comprised 81 PDAC samples divided into treatment groups: 7 untreated cases, 8 neoadjuvant FOLFIRINOX cases, 4 neoadjuvant gemcitabine + nab-paclitaxel cases, 1 mixed treatment case (FOLFIRINOX and gemcitabine + nab-paclitaxel), and 1 chemo-radiotherapy (Chemo-RT) case. Tumor samples were spatially sampled 2–4 times per tumor, followed by histology, imaging, multi-omics analysis, and bulk RNA sequencing. Single-cell RNA sequencing (scRNA-seq) data were generated for all 73 samples. After quality control, cells with a mitochondrial RNA expression greater than the 75th percentile were excluded, resulting in a total of 232,764 cells retained for downstream analysis.

**Text embedding of gene description and path connection from PubMedBert**

We embedded 33,008 genes in the dataset using the large language model PubMedBert18. First, we crawled the descriptions of all the genes, including those from the four datasets NCBI Gene, GeneCards, UniProtKB/Swiss-Prot, and Tocris. Secondly, we input them into PubMedBert 18 to generate 768-dimensional gene embedding. We obtained a total of 22,173 valid gene embeddings.

In addition, we screened these genes through the network to generate gene pairs with biological connections, and converted them into text, such as "In this pathway: receptor gene ABCA1 connect to target gene DPPA2." We sent the converted text to PubMedBert 18 to generate Path-Embedding.

**Cell Annotation**

We used principal component analysis (PCA) to reduce the dimensionality of the raw data. We projected the gene data to 50 dimensions and used the uniform manifold approximation and projection (UMAP) method using the first 25 principal components (PCs). This approach ultimately facilitated the visualization of the data in two-dimensional space (Figure 1A). Using the marker genes in these reference datasets, we identified and labelled 14 cell types with significant expression, including Endothelial, general CAF, CD4 T cells, cDCs, Macrophages, NK, CD8 T cells, B cells, Plasma, Islet, Tumor, Treg, Fibroblast and Acinar (Figure 1B).

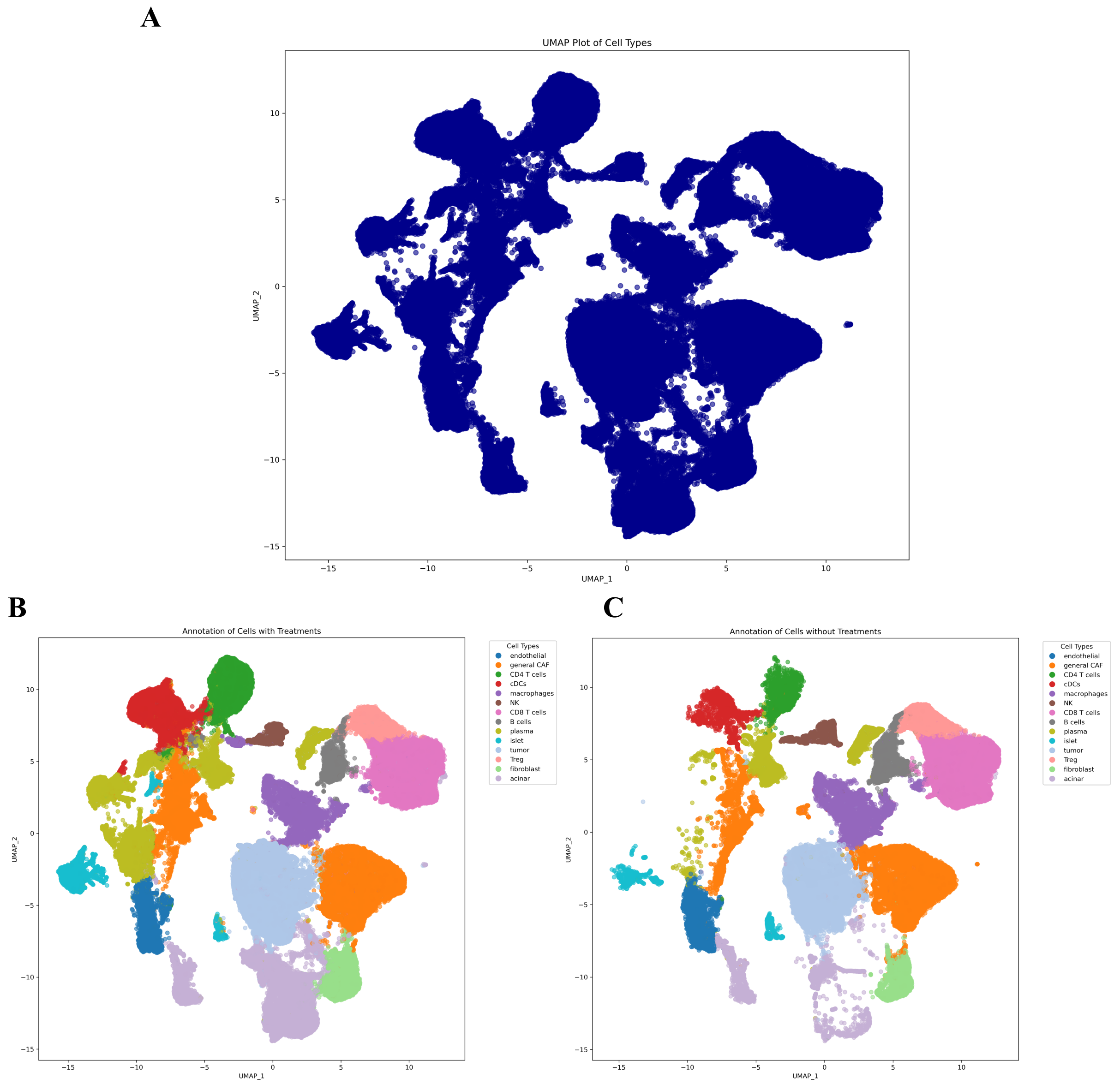

**Figure 1**. Dimension reduction and clustering results using sc2MeNetDrug. **(A)** Dimension reduction result using 20 PCs of human PDAC dataset. Clustering result with a UMAP resolution of 0.01, and 20 PCs of **(B)** Patients with treatments **(C)** Patients with no treatment.

**Cell Distribution**

The distribution of cell types showed significant differences of PDAC patients with treatments and without treatment (Figure 2A).

**Immunosuppression Enhancement: Significant Reduction in T Cells**

Compared with the “without treatment” group, the “treatment” group had a significant decrease in CD8+ T cells and Treg cells. Specifically, CD8+ T cells decreased from 24.9% to 8.1%, while Treg cells decreased from 3.8% to 2%. This observation is consistent with the immunosuppressive nature of PDAC, which lacks immune cells even in their presence, as they often exhibit pro-tumorigenic features. The expansion of CD8+ T cells further supports systemic suppression in PDAC 19. Furthermore, immunosuppressed systemic regions provide a cozy haven for disseminated tumor cells, thereby promoting disease recurrence20 .The decrease in CD8+ T cells in the “treatment” group further supports the notion that treatment may exacerbate immunosuppression in the tumor microenvironment. Furthermore, Tregs are the most abundant T cell population in PDAC and play a key role in maintaining the immunosuppressive microenvironment. The observed decrease in Treg cells from 3.8% to 2% after treatment may reflect immunoregulatory mechanisms affected by Tregs, as there are increasing reports showing a paradoxical association between tumor infiltration of Tregs and improved patient outcomes 21. These findings suggest that treatment-induced reductions in immune effector cells such as Tregs and CD8+ T cells may enhance the immunosuppressive environment in PDAC.

**Antigen Presentation and Tumor Microenvironment Remodeling: Significant Increase in cDCs and Acinar Cells**

A significant increase in cDCs (from 1.6% to 9%) and acinar cells (from 3.3% to 12%) was observed in the "with treatment" group compared to the "without treatment" group. The rise in cDCs indicates enhanced antigen presentation activity, which could be a result of the treatment altering the tumor microenvironment to recruit or activate these dendritic cells. Similarly, the increase in acinar cells might represent changes in stromal cell populations or reprogramming of the tumor microenvironment during treatment. Studies have reported an increase in antigen-presenting cells following immunomodulatory treatments, suggesting their potential role in

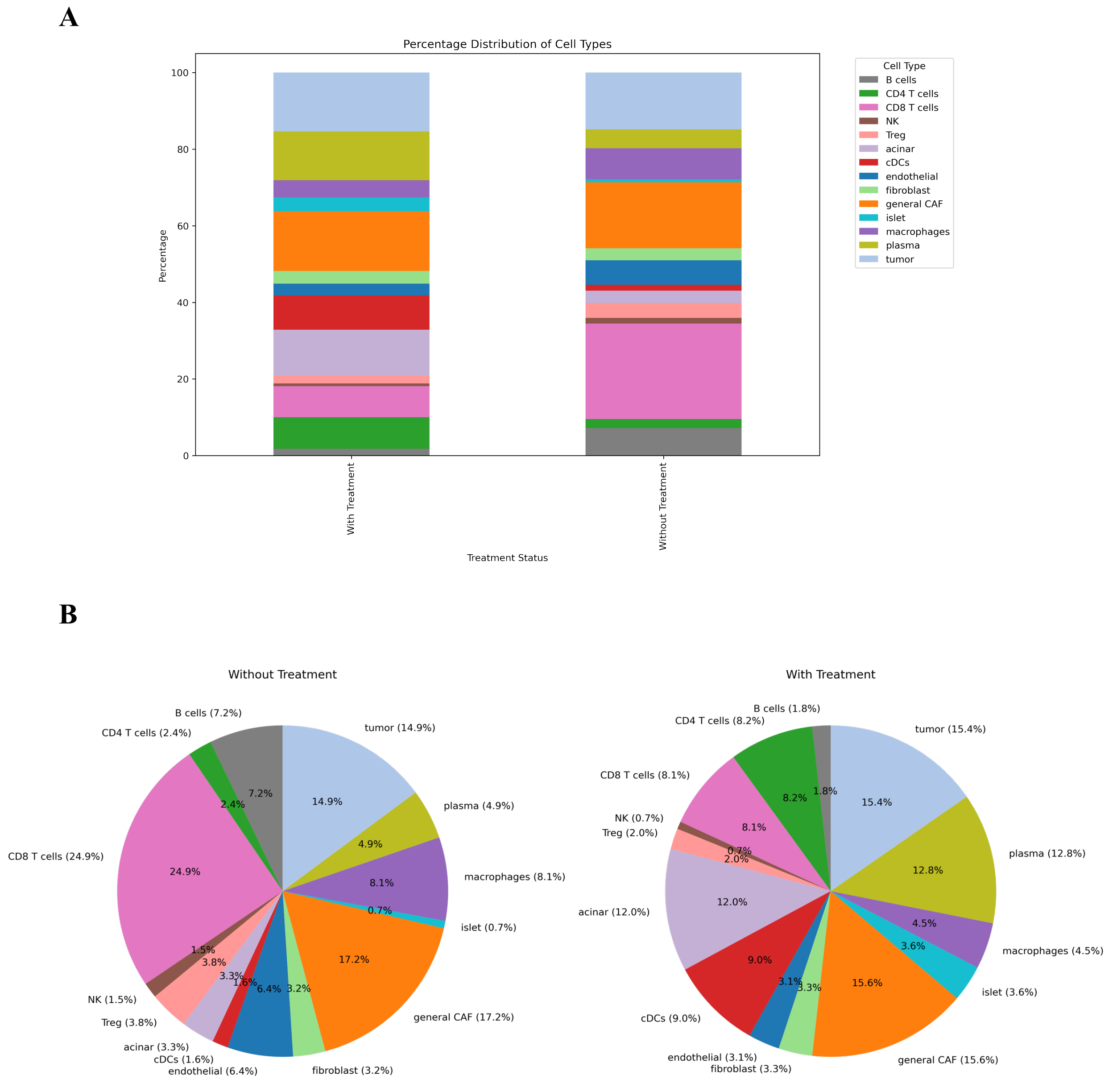

**Figure 2**. **(A)** Cell types found between groups of patients with treatments and without treatments. **(B) (C)** Percentage of cell types found in group of patients with treatments and without treatments

remodeling the tumor microenvironment to enhance therapeutic outcomes.

**Cytokine and Immune Regulation Changes: Increase in Plasma Cells and Helper T Cells**

An increase in plasma cells (from 4.9% to 12.8%) and CD4+ T cells (from 2.4% to 8.2%) was observed in the "with treatment" group compared to the "without treatment" group. Plasma cells are known to secrete antibodies, which can modulate immune responses within the tumor microenvironment. The increase in CD4+ T cells suggests a shift in the immune balance, potentially favoring immune regulation or activation of specific immune pathways. Such changes might indicate that treatment induces a compensatory mechanism to regulate the immune response or mitigate inflammation. Previous studies have demonstrated that increased helper T cells and plasma cells correlate with better immune regulation and could enhance therapeutic efficacy.

## Results

### Model evaluation and comparison

To evaluate the classification performance of our model, we applied it to the human cirrhosis dataset of Endothelial cells, Macrophages, and T cells separately to predict each cell's Healthy/Cirrhosis status. For each cell type, we repeated the training three times, each time randomly splitting the entire dataset into a training subset, a validation subset, and a test subset at a ratio of 0.7/0.1/0.2. We report the average performance and standard deviation of the test set over all three runs. The detailed experimental setup is described in the Methods section. The detailed results are shown in Table 1. We also applied it to the Alzheimer's disease (AD) scRNA-seq dataset (GSE164507) collected from the Gene Expression Omnibus (GEO) database (Wang et al., 2021). We specifically selected three cell types from the TAFE4 group for analysis: excitatory neurons (Ex), microglia (Mic), and astrocytes (Ast). The results are summarized in Table 2.

| | Accuracy | Recall | Precision | Specificity | F1 | AUC |
|---|---|---|---|---|---|---|
| Endo | 00.89 ± 0.02 | 0.96 ± 0.03 | 0.78 ± 0.04 | 0.85 ± 0.09 | 0.87 ± 0.03 | 0.96 ± 0.03 |
| Mac | 00.88 ± 0.03 | 0.95 ± 0.02 | 0.75 ± 0.02 | 0.84 ± 0.07 | 0.85 ± 0.04 | 0.95 ± 0.03 |
| T-cell | 00.86 ± 0.02 | 0.93 ± 0.03 | 0.72 ± 0.04 | 0.82 ± 0.04 | 0.87 ± 0.03 | 0.94 ± 0.02 |

**Table 1**. Evaluation results on human cirrhosis dataset

| | Accuracy | Recall | Precision | Specificity | F1 | AUC |
|---|---|---|---|---|---|---|
| Excitatory neurons | 00.84 ± 0.01 | 0.88 ± 0.03 | 0.83 ± 0.02 | 0.81 ± 0.05 | 0.85 ± 0.01 | 0.91 ± 0.01 |
| Microglia | 00.88 ± 0.01 | 0.93 ± 0.02 | 0.82 ± 0.02 | 0.76 ± 0.04 | 0.87 ± 0.01 | 0.89 ± 0.01 |
| Astrocytes | 00.83 ± 0.04 | 0.92 ± 0.01 | 0.82 ± 0.04 | 0.73 ± 0.07 | 0.86 ± 0.04 | 0.83 ± 0.04 |

**Table 2**. Evaluation results on AD dataset

We compare our work with random forests, multilayer perceptron, convolutional networks, and our models from previous phases. We find that our model significantly outperforms all other models, including the model we developed in the previous phase. It is worth mentioning that at this stage of our research, we introduced gene embedding and path embedding from the output of the large language model (PubMebBert) as part of our model training, understanding, and applying the information that extracts from the large language model has also become key to the progress of the model. Table 3,4 lists the complete results we achieved in the following tasks. All reported data are from the test set.

| Model | Accuracy |
|---|---|
| Random | 0.518 |
| MLP | 0.567 |
| Random Forest | 0.585 |
| Graph Convolutional Network | 0.721 |
| Pathfinder (our previous model) | 0.813 |
| Our current model | **0.882 ± 0.02** |

**Table 3.** Accuracy comparison results on human cirrhosis dataset

| **Model** | **Accuracy** |
|---|---|
| Random | 0.501 |
| MLP | 0.542 |
| Random Forest | 0.587 |
| Graph Convolutional Network | 0.692 |
| Pathfinder (our previous model) | 0.754 |
| Our current model | **0.847 ± 0.05** |

**Table 4.** Accuracy comparison results on AD dataset

**Prediction Tasks of signaling pathway inference**

We used the golden network proposed by McCalla et al. 24.to explore the rationality of the pathways we generated. The study by McCalla et al. collected multiple experimental regulatory interaction networks from public databases and literature as gold standards for network inference algorithms24. These experiments are based on ChIP-chip, ChIP-seq, or regulatory factor perturbation after global transcriptome analysis. They obtained multiple networks based on ChIP and transcription factor perturbation experiments for each organism and cell type, which are called "ChIP" and "Perturb". For human ESC (hESC) cell lines, they used CIS-BP 25, ENCODE26, and JASPAR 27 databases for construction. We used their “hESC_ChIP” and “hESC_Perturb” gold standard networks to test the results obtained by our model on the human Cirrhosis and human AD datasets.

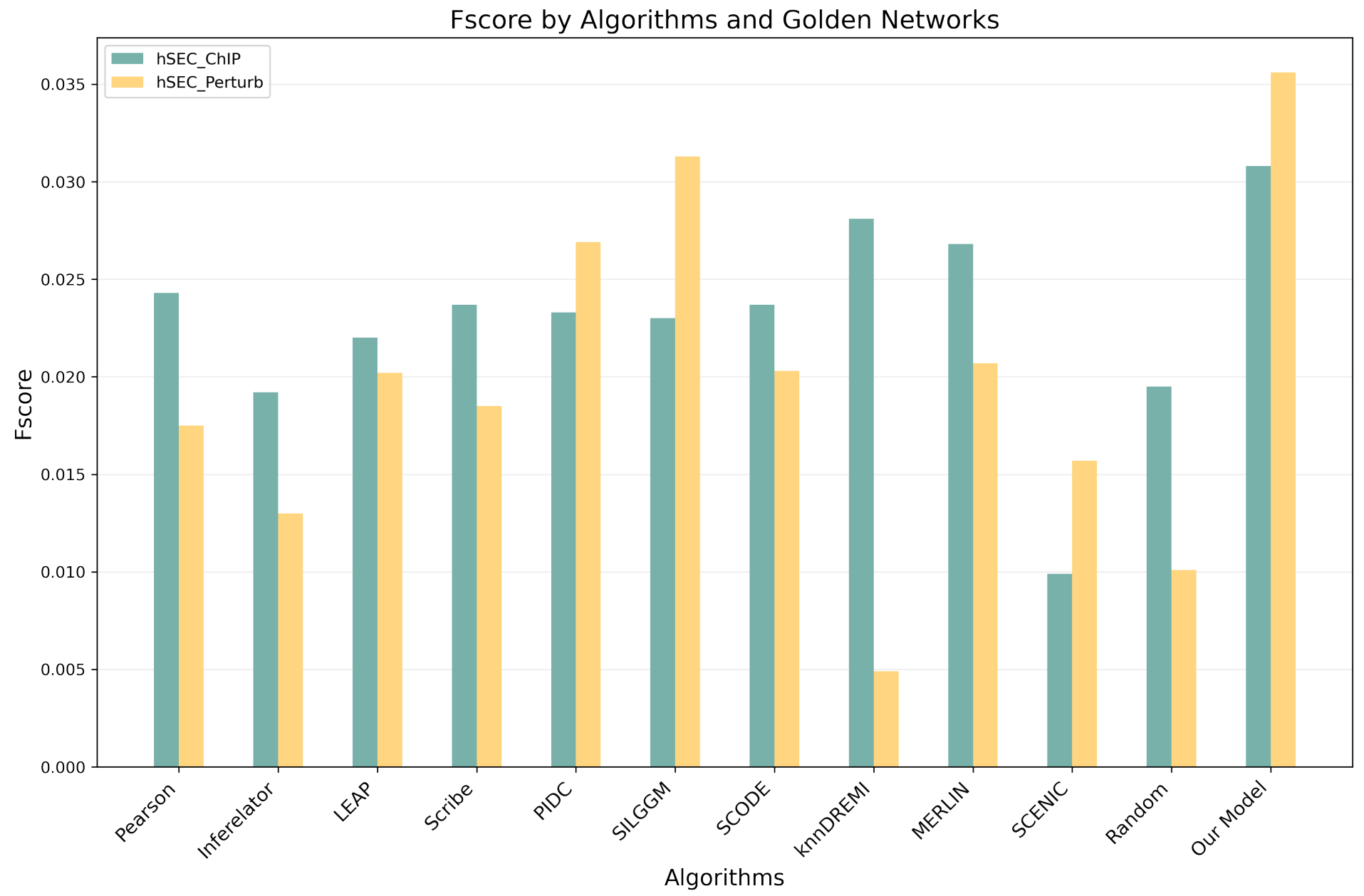

**Figure 6.** F-score comparison results on human cirrhosis dataset.

To evaluate the accuracy of the inferred network, we used F-score. Before calculating each accuracy metric, we filtered the inferred network and the gold standard network to only include edges containing transcription factors and target genes from the intersection of the gold standard gene set and the original expression matrix gene set. This is done to eliminate any penalty for edges included in the gold standard that cannot be inferred by the algorithm due to missing expressions in the original data24. In addition, due to the different output sizes between different networks, we limited the gene pairs involved in the test to the smallest output gene pairs between different model networks. The F-score is used to measure the accuracy of the most confident edge in the inferred network. It is done by sorting the inferred edges in descending order of confidence, calculating the precision and recall when the edges are included one by one relative to the gold standard, and finally reporting the area under this curve. When comparing the network output by our model with the network output by the other 11 models, our output achieved the highest F-score under both " hESC_ChIP" and " hESC_Perturb" gold standards. The detailed results are shown in Figure 6.

## Summary and conclusion

In conclusion, the TNGs and joint LLM-GNN models are important approaches for scientific discovery. The design and implementation of novel and efficient LLM-GNN and TNGs models are a potentially new field for scientific discovery in the age of AI.

## Acknowledgments

This study was partially supported by NIA R56AG065352, NIA 1R21AG078799-01A1, NINDS 1RM1NS132962-01, Children’s Discovery Institute (CDI) M-II-2019-802, and NLM 1R01LM013902-01A1. The results published here are in whole or in part based on data obtained from the AD Knowledge Portal (https://adknowledgeportal.org).

**Supp results: Disease biomarkers mining and pathway analysis**

**Our analysis also identified new disease biomarkers and signaling pathways.**

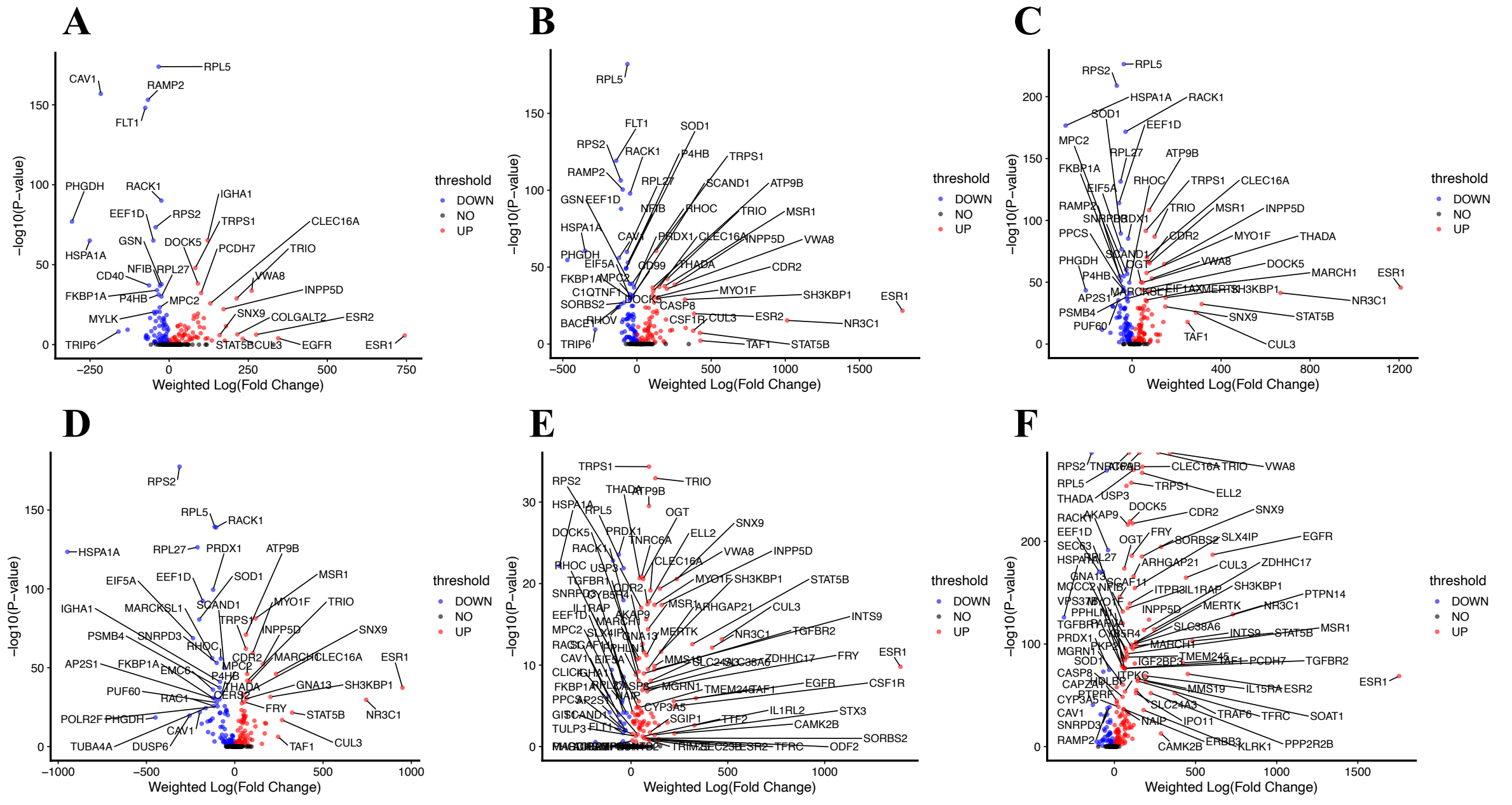

**Figure 3**. Top expression changes of genes from without treatment state to with treatment state. **(A)** Acinar. **(B)** Fibroblast. **(C)**CAF. **(D)** Islet. **(E)** Macrophages **(F)** Tumor

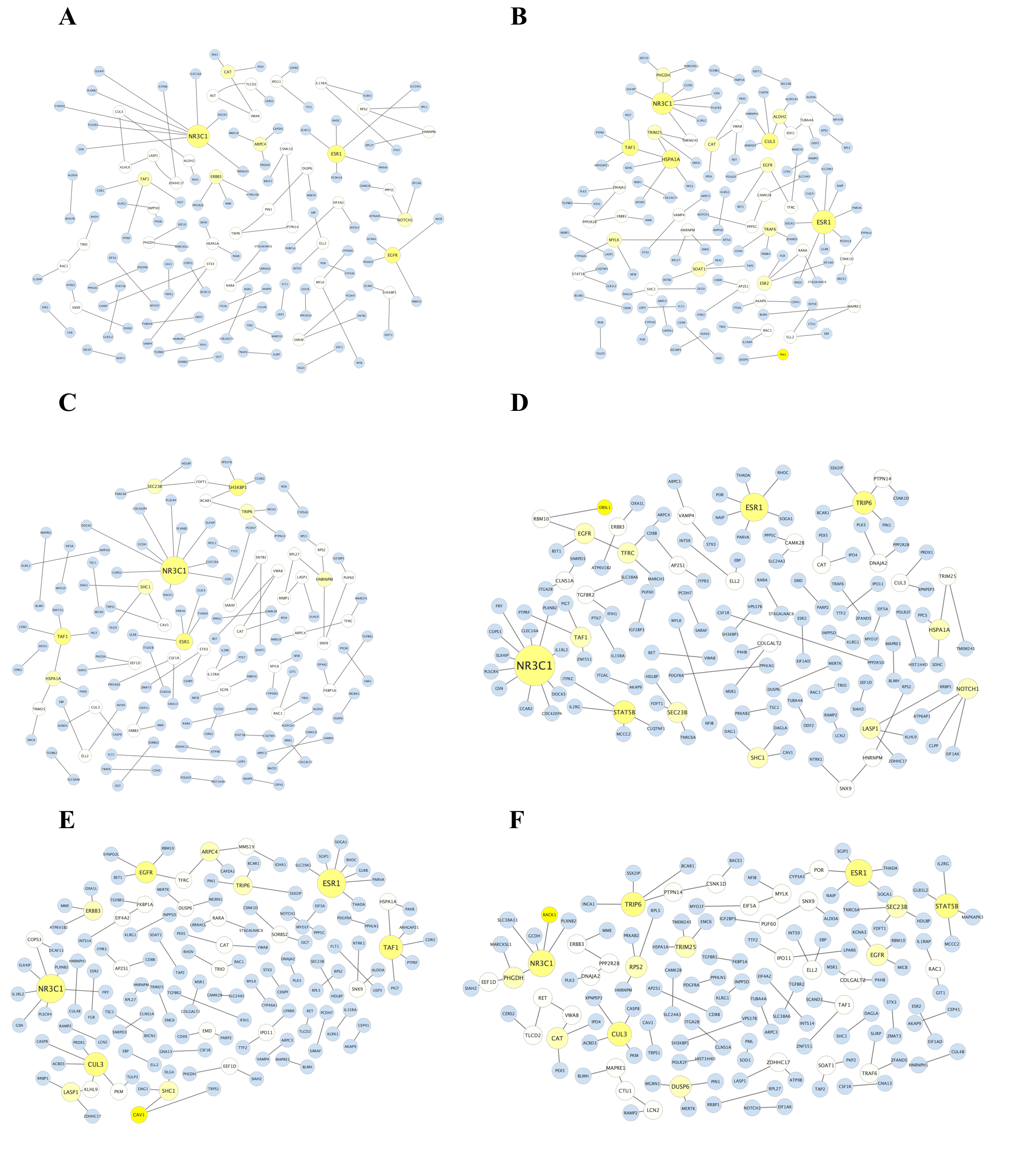

**Figure 4**. Important intra-cell signaling pathways at **(A)** CAF. **(B)** Fibroblast. **(C)** Tumor. **(D)** Acinar. **(E)** Islet. **(F)** Macrophage.

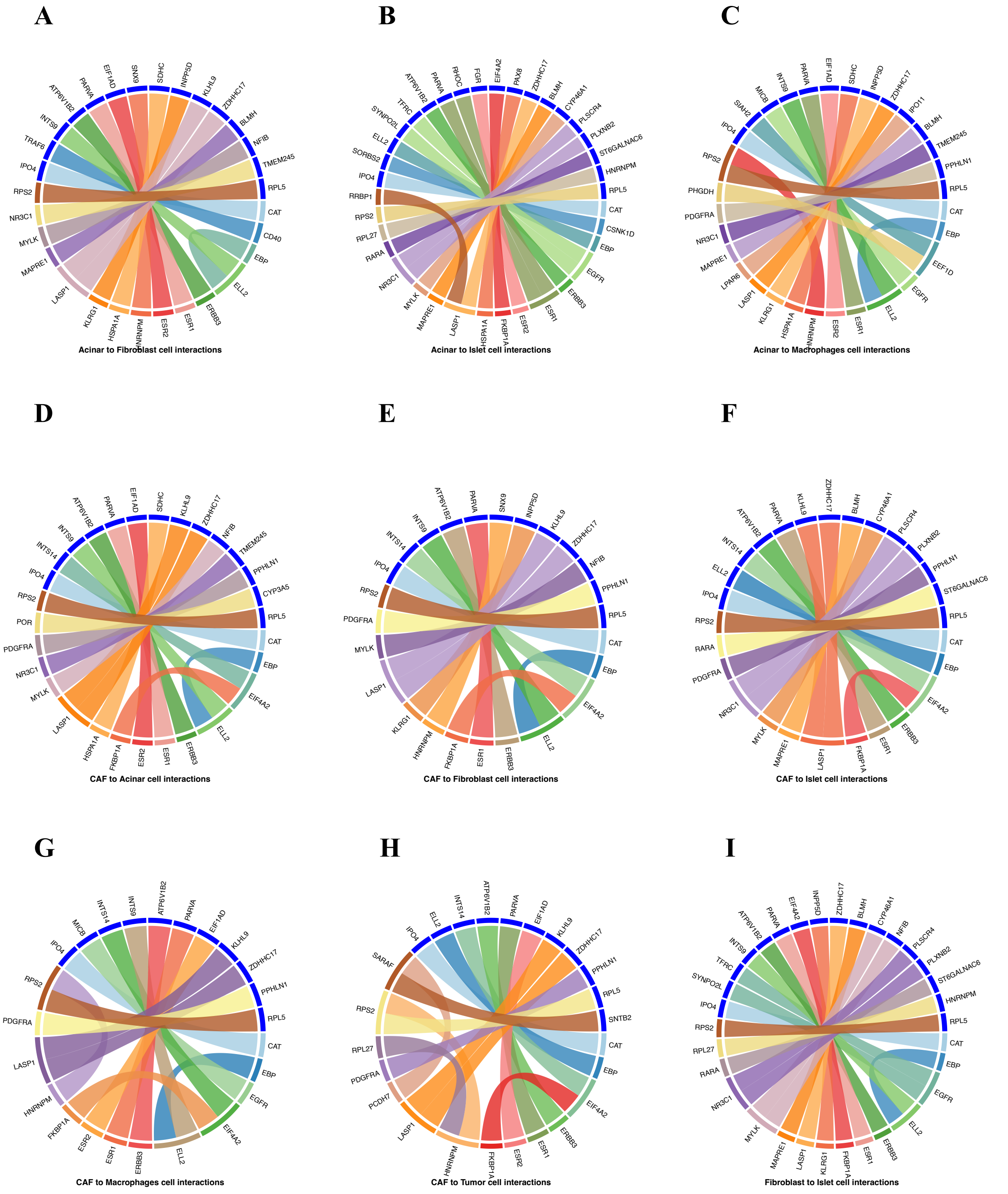

**Figure 5**. Inter-cell pathways. **(A)** Acinar - Fibroblast. **(B)** Acinar - Islet. **(C)** Acinar - Macrophages. **(D)**CAF - Acinar. **(E)** CAF - Fibroblast. **(F)** CAF - Islet. **(G)** CAF - Macrophages. **(H)** CAF - Tumor. **(I)** Fibroblast - Islet.

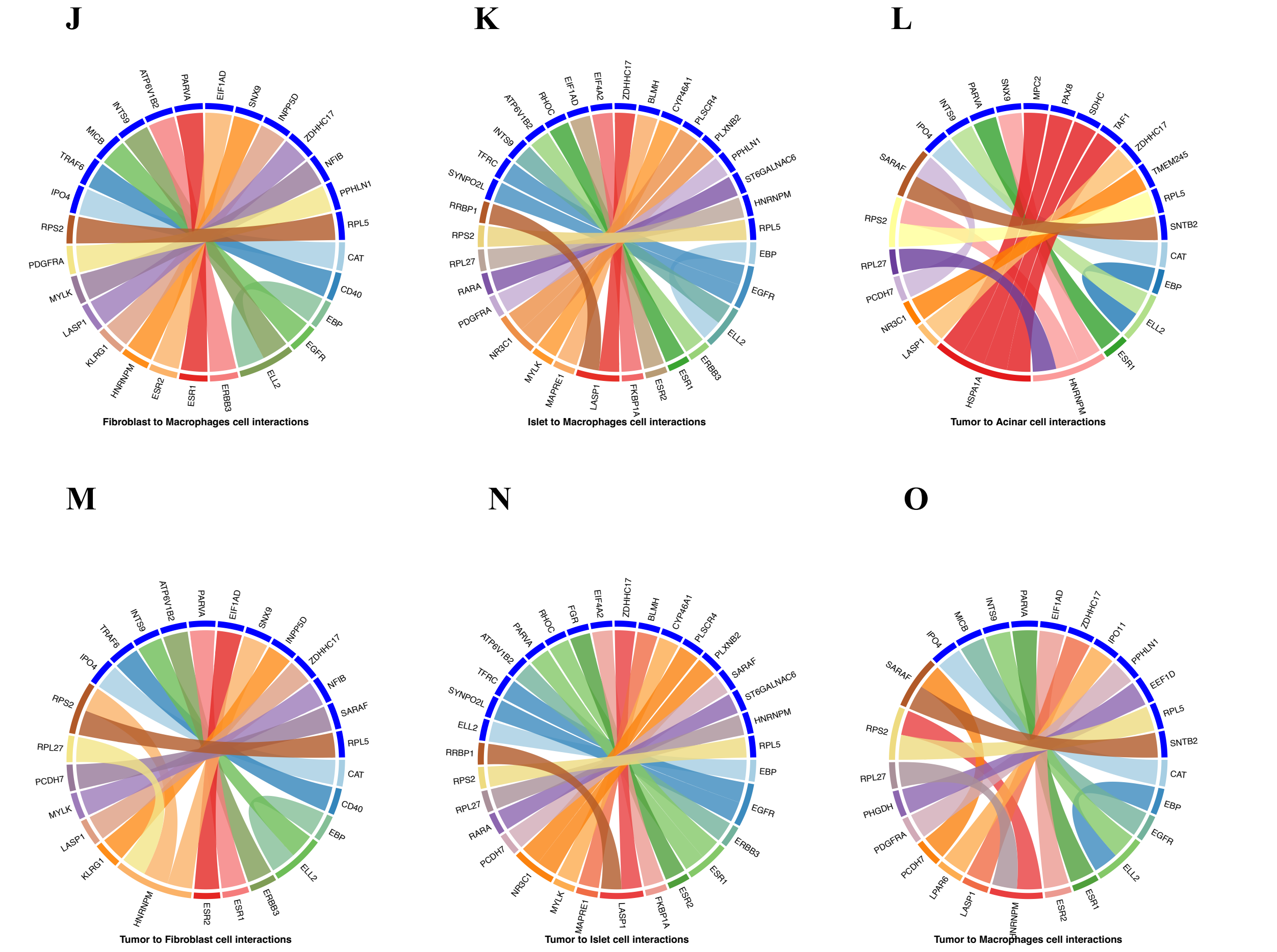

**Figure 5.** Inter -cell pathways. **(J)** Fibroblast - Macrophages. **(K)** Islet -Macrophages. **(L)** Tumor - Acinar. **(M)**Tumor - Fiborblast. **(N)** Tumor - Islet. **(O)** Tumor -Macrophages.